\documentclass[conference,a4paper]{IEEEtran}

\usepackage[T1]{fontenc}
\usepackage[utf8]{inputenc}

\usepackage{cite}
\usepackage[cmex10]{amsmath}
\usepackage{amssymb,amsfonts}
\usepackage{array}
\usepackage[caption=false,font=footnotesize]{subfig}
\usepackage{dblfloatfix}

\usepackage{algorithmic}
\usepackage{graphicx}
\graphicspath{{graphics/}{moregraphics/}}
\DeclareGraphicsExtensions{.pdf,.PDF,.jpeg,.JPEG,.jpg,.JPEG,.png,.PNG}

\usepackage{booktabs}

\usepackage{siunitx}

\usepackage{textcomp}
\usepackage{xcolor}
\usepackage[spaces,hyphens]{url}

\def\BibTeX{{\rm B\kern-.05em{\sc i\kern-.025em b}\kern-.08em
    T\kern-.1667em\lower.7ex\hbox{E}\kern-.125emX}}

\begin{document}

\title{Understanding Consumer Preferences for Movie Trailers
from EEG using Machine Learning}

\author{\IEEEauthorblockN{Pankaj Pandey}
\IEEEauthorblockA{Computer Science and Engineering\\
\textit{Indian Institute of Technology Gandhinagar}\\
Gujarat, India\\
pankaj.p@iitgn.ac.in}
\and
\IEEEauthorblockN{Raunak Swarnkar}
\IEEEauthorblockA{Centre for Cognitive \& Brain Sciences \\
\textit{Indian Institute of Technology Gandhinagar}\\
Gujarat, India \\
raunak.swarnkar@iitgn.ac.in}\and
\IEEEauthorblockN{Shobhit Kakaria }
\IEEEauthorblockA{Centre for Cognitive and Brain Sciences \\
\textit{Indian Institute of Technology Gandhinagar}\\
Gujarat, India \\
shobhit.kakaria@iitgn.ac.in}
\and
\IEEEauthorblockN{Krishna Prasad Miyapuram}
\IEEEauthorblockA{ Centre for Cognitive and Brain Sciences\\Computer Science and Engineering\\
\textit{Indian Institute of Technology Gandhinagar}\\
Gujarat, India \\
kprasad@iitgn.ac.in}
}

\maketitle
\begin{abstract}Neuromarketing aims to understand consumer behavior
using neuroscience. Brain imaging tools such as EEG
have been used to better understand consumer behavior that goes beyond self-report measures which can be a more accurate measure to understand how and why consumers prefer choosing one product over another. Previous studies have shown that consumer preferences can be
effectively predicted by understanding changes in evoked
responses as captured by EEG. However, understanding
ordered preference of choices was not studied earlier. In
this study, we try to decipher the evoked responses using
EEG while participants were presented with naturalistic
stimuli i.e. movie trailers. Using Machine Learning techniques to mine the patterns in EEG signals, we predicted
the movie rating with more than above-chance, 72\% accuracy. Our research shows that neural correlates can be
an effective predictor of consumer choices and can significantly enhance our understanding of consumer behavior.

\end{abstract}

\begin{IEEEkeywords}
Neuromarketing, EEG, Machine Learning, Discrete Wavelet Decomposition
\end{IEEEkeywords}

\section*{Introduction}
Consumer neuroscience systematically aims to understand
consumer behavior and underlying preferences through the
lens of psychology, neuroscience, marketing and economics
\cite{ariely2010neuromarketing}. Currently, consumer research relies
on the stimulus-reponse model to capture the underlying brain
processes. The development of human neuroscience tools to
understand the latent mechanisms gave rise to progress of
using neuroscience to effectively understand what the consumer prefers going beyond the traditional subjective scores\cite{hsu2015neuroscience,silberstein2008brain}. Understanding what contributes to the observed behavior by complex,
naturalistic stimuli has immense importance in marketing research \cite{hubert2008current,dmochowski2014audience,plassmann2007can}. Preferences of consumers have been widely studied using other
modes than self-reported measures. The idea of ’preference’
has been historically approached differently in psychology, economics, marketing and neuroscience. One neuroscience study
has shown that behavior preferences can be attributed to brain
activity in the ventromedial prefronal (vmPFC) cortex and
ventral striatum \cite{mcclure2004neural}.

Electroencephalography (EEG) is being recently used as a
tool to understand the neural correlates of consumer behavior due to its advantage of providing temporal resolution of
stimulus-response . Previous research suggests using EEG to understand the patterns of brain activity of participants
watching advertisements to understand their buying preferences \cite{wang2008validity}. The main motivation behind
this current study is to understand consumer neuroeconomics
from a machine learning perspective, primarily to understand
movie preferences, which is a line of research that has not
received much attention. For instance, the US movie box office revenue is above \$10 Billion per year. However, invest-
ing in such projects is risky. It has been found that only
36\% of movies achieve a break-even over the production cost,
in terms of profits. Movie producers spend huge amounts
on movie trailers and use different marketing techniques to
gauge consumer reactions, and eventually, box-office earnings.
Study used EEG for predicting brain responses
to movie trailers based on individual consumer preferences,
where brainwave signals in the beta frequency oscillation range
were found to be a good predictor for like/dislike of a movie
trailer \cite{boksem2015brain}. In another work, EEG was
used to study consumer behavior where it was found that future choices in terms of which product a consumer will prefer
buying, was predicted using EEG \cite{kang2015investigation}.

In this study, we study consumer behavior in terms of movie
trailer preferences using both neural and behavioral measures,
with the main aim to use pattern recognition techniques to
understand evoked responses in EEG. The present study tries
to systematically investigate whether neural correlates can be
a valuable predictor of individual choice preferences.

\section*{Materials and Methods}

\subsection{Experiment Design}
The dataset was collected using a 128 channel Net Station
Electroencephalography (EEG) device from 18 healthy
individuals (13 were men, and 5 were women) at Indian
Institute of Technology, Gandhinagar. The mean age of
participants was 22.4 years within the range of 18-26 years.
No participant reported any history of neurological or
psychiatric disorder. Fifteen participants were right-handed,
and three were left-handed. Data acquisition was done using
EGI Netstation 5.2. The experiment consisted of 12 trial
blocks. In the experiment, every subject was asked to watch
12 trailers of upcoming movies, and after each trailer, there
were four questions to be answered on a likert scale of 1-5:

\begin{enumerate}
    \item Rating
    \item Familiarity
    \item Purchase Intent
    \item Willigness to Spend
\end{enumerate}

\begin{figure}
  \centering
  \includegraphics[width=\linewidth]{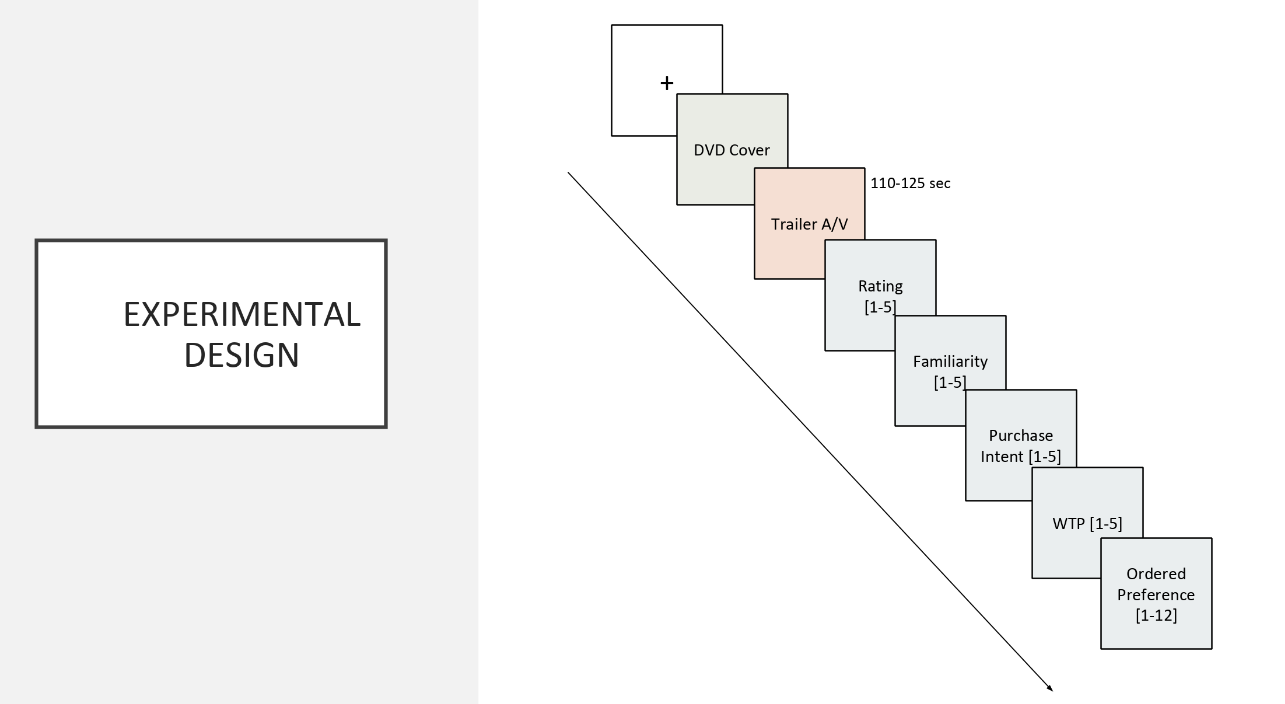}
  \caption{Experiment design}
  \label{ed}
\end{figure}

Lastly, participants were asked to report an ordered pref-
erence to arrange the movies in descending order, the movie
trailers they liked the most to least preferred. Experiment Design is shown in fig. \ref{ed}.

This step consisted of removal of data points in the dataset
that were erroneous. Those samples where the electrodes were
not in proper location, or not in contact with the scalp, were
excluded from the dataset before further processing. EEG signal is susceptible to noise or artefacts and has to be corrected
as part of the signal pre-processing step as mentioned in the
below. All the pre-processing was performed using EEGLab
Package in MATLAB \cite{delorme2004eeglab}. Further,
bad channels were rejected using Artifact Subspace Reconstruction using Makato’s Pre-processing pipeline \cite{makoto}. Re-referencing was performed across all channels. The following artefacts were identified and removed.

\subsubsection{Physiological Artefacts}
Electrooculogram (EOG): This is the electrical noise gener-
ated by eye blink and cornea movement that is captured in
the EEG signal and has to be removed. It can be estimated
as the change of potential in electrodes near the eyes at Fp1-
Fp2 (Fronto Parietal). Fluttering of the eyelids appears as a
3Hz –10Hz signal, and hence was removed using Band Pass
filters.

Electromyogram (EMG): This is the electrical “noise” gener-
ated by muscle activity. Facial Muscle movement; swallowing,
grimacing, chewing can be captured in EEG and has to be
removed. This noise commonly appears in the frontal and
temporal electrodes.

\subsubsection{External Artefacts}
Physical movement: This can lead to lose contact of electrode
due to abrupt physical movement of subjects and is captured
as a high amplitude, low frequency noise.
Electrode Contact: Poor electrode contact gives rise to low
frequency artifacts and all such trials were removed which had
lost any electrode contacts.

\subsection{Feature Extraction}

In order to extract features from EEG signals, a well-
established method is to decompose the mother wavelet into
sub-frequency bands. There are primarily 5 types of frequency
bands as shown in table \ref{fb}.

\begin{table}
  \begin{center}
    \caption{Frequency Bands}
    \label{fb}
    \begin{tabular}{l|c} 
    \hline
      \textbf{Frequency Band} & \textbf{Frequency Range(Hz)}\\
      \hline
      Delta & 0-3.5\\
      \hline
      Theta  & 4-7\\
      \hline
      Alpha & 8-13\\
      \hline
      Beta & 14-30\\
      \hline
      Gamma & 30-60\\
      \hline
    \end{tabular}
  \end{center}
\end{table}

A general method for decomposition is to use Fast Fourier
Transform. But since EEG electrode signal is non-stationary,
FFT might not be the best alternative \cite{gross2014analytical,azim2010feature,hamad2016feature}. Instead, we used Discrete
Wavelet Transform (DWT) Method (db-8 wavelet) using Mat-
lab Wavelet Toolbox to extract two features, Power and En-
tropy for all 5 frequency sub-bands in range of 0-60 Hz, across
all 128 channels.

\subsection{Feature Elimination}
Since the feature set had a large number of features consisting
of 5 DWT features for both Power and Entropy, two feature
elimination techniques were used:
\begin{enumerate}
\item Recusrive Feature Elimination: This algorithm ranks
the features by associating the weights with features and
prune the features as per the weights. It forms the smaller set
after each iteration and terminate until the given (k) number
of features is achieved.
\item Sequential Backward Selection (SBS): A greedy search technique
to reduce dimensions of the feature vector from a d space to
lower dimensional k space.\end{enumerate}
These two methods were used to select features which were
further trained with the set of classifiers.

\subsection{Machine Learning Classifiers}
Choosing the correct set of Machine Learning models is a
crucial step for classification. 9 Machine Learning classifiers
were used to predict the labelled class using the extracted
feature set, which includes: k Nearest Neighbors, Random
Forest, Quadratic Discriminant Analysis, Decision Tree, Multilayer Perceptron, Gaussian Naive Bayes, Gaussian Process
Classifier, Ridge Classifier, Support Vector Classifier. Every
classifier was trained with a Cross-Validation set and test for
performance using the test set. The hyper-parameters of the
classifiers were tuned iteratively. We trained the models using python. \cite{pedregosa2011scikit, raschkas_2018_mlxtend}.

\subsection{Results}
The total sample size of 216 data points was divided into train set with 151 and test set with 65 samples. All the 4 bands were found to have discriminating information for the classifier. Machine learning classifiers were then used to train and were tested accordingly, with the top 5 classifiers using 10 Fold Cross-Validation with their results as shown in table \ref{cp}

\begin{table}
  \begin{center}
    \caption{Classification Performance}
    \label{cp}
    \begin{tabular}{l|c|r}
      \textbf{Classifier} & \textbf{Feature Elimination} & \textbf{Test Accuracy}\\ 
      \hline
      kNN & RFE & 0.7237\\ 
      \hline
      Random Forest & SBS & 0.7069\\ 
      \hline
      kNN & SBS & 0.6923\\
      \hline
      Multi-layer Perceptron & SBS & 0.6769\\
      \hline
    \end{tabular}
  \end{center}
\end{table}

\section{Discussion}
The use of EEG in Neuromarketing to understand consumer
behavior is an important area of research given its wide implications. Machine learning techniques have aided immensely in
this regard, to decode the information in EEG signals. In this
study, we attempted to find whether consumer preferences can
be predicted using EEG signals and achieved high accuracy to
predict ratings. The main aim to predict ordered preference
of movie trailer still needs to be studied, which is the primary
goal of this study.
However, using EEG for consumer research has its own
challenges. EEG signals have a low Signal-to-Noise ratio and
hence, it becomes a challenge to accurately process the signal. Sensitivity to various artefacts also poses problems in
the data cleaning process. Better signal processing techniques
could further improve the Signal-to-Noise ratio and could improve classification performance which needs to be explored.
However, the most significant challenge faced was the small n-large-p problem i.e. large number of features as compared
to number of samples. One limitation of our work is that we
had averaged the feature values across channels, which may
have led to loss of information. One way to resolve this would
be to use dimensionality reduction techniques. We made a
preliminary attempt to use PCA but did not get any significant improvements. Using non-linear dimensionality reduction techniques such as t-SNE and UMAP which are considered to be the state-of-the-art techniques need to be further
explored. Deep Learning techniques have been found to be
quite powerful when learning internal representations of features, which could be significant as a dimensionality reduction
step, which can be further explored.

\bibliographystyle{IEEEtran}
\bibliography{references}

\end{document}